\documentclass[10pt, a4paper]{article}

\usepackage{url}

\usepackage{breakurl}
\usepackage[breaklinks]{hyperref}

\usepackage{lrec-coling2024} 
\usepackage{rotating}
\usepackage{makecell}
\usepackage{booktabs}

\usepackage{graphics}
\usepackage{booktabs}
\usepackage{textcomp}
\usepackage{enumitem,xcolor}
\title{\textsc{DORE}: A Dataset For Portuguese Definition Generation\\ \vspace*{.5\baselineskip}}

\name{Anna Beatriz Dimas Furtado\textsuperscript{1}, Tharindu Ranasinghe\textsuperscript{2}, \\ \large \textbf{Frédéric Blain\textsuperscript{3}, Ruslan Mitkov\textsuperscript{4}}} 

\address{\textsuperscript{1}University of Galway, IE, \textsuperscript{2}Aston University, UK, \\ \textsuperscript{3}Tilburg University, NL, \textsuperscript{4}Lancaster University, UK \\
         \texttt{annabeatriz.dimasfurtado@universityofgalway.ie}, \texttt{t.ranasinghe@aston.ac.uk}, \\ \texttt{f.l.g.blain@tilburguniversity.edu}, \texttt{r.mitkov@lancaster.ac.uk} \\
         }

\abstract{
Definition modelling (DM) is the task of automatically generating a dictionary definition for a specific word. Computational systems that are capable of DM can have numerous applications benefiting a wide range of audiences. As DM is considered a supervised natural language generation problem, these systems require large annotated datasets to train the machine learning (ML) models. Several DM datasets have been released for English and other high-resource languages. While Portuguese is considered a mid/high-resource language in most natural language processing tasks and is spoken by more than 200 million native speakers, there is no DM dataset available for Portuguese. In this research, we fill this gap by introducing \textsc{DORE}; the first dataset for \textbf{D}efinition M\textbf{O}delling for Po\textbf{R}tugu\textbf{E}se containing more than 100,000 definitions. We also evaluate several deep learning based DM models on \textsc{DORE} and report the results. The dataset and the findings of this paper will facilitate research and study of Portuguese in
wider contexts.
 \\ \newline  }

\begin{document}

\maketitleabstract

\section{Introduction}
\label{sec:intro}

Definitions play a key role in the globalised world; they are useful for a wide range of audiences, from professionals to students \cite{dziemianko-2020-dict-usefulness}. They are also the building blocks of effective communication and understanding within the Information Society; it is imperative to have domain experts to ensure their accuracy, clarity, and coherence, which can be expensive \cite{sanmartin-2023-termdefs}. Yet, crafting high-quality definitions demands time and effort due to their intricate and complex nature \cite{vazquez-gouws-2023}. Given the array of challenges involved, the manual creation of definitions proves to be a difficult, expensive, and arduous task \cite{sanmartin-2023-termdefs}.

Introduced by \citet{noraset2017} as a supervised machine learning (ML) task \cite{ni-wang-2017-learning}, definition modelling (DM) addresses these challenges by designing systems capable of automatically generating definitions for a specific word. Beyond its immediate application of generating definitions for dictionaries, DM can be useful for completing the WordNet, providing resources for language learners, language preservation and language description. Furthermore, it has been used as a window in explainable AI to shed the light on the quality of embeddings \cite{mickus2022codwoe}, besides to provide results for studying LLM hallucinations and overgeneration mistakes \footnote{As proposed by Zosa et al. in \href{}{https://helsinki-nlp.github.io/shroom/}}. It has also been suggested as a way to detach word-sense disambiguation from word inventories \cite{bevilacqua2020generationary}. 

Most studies consider DM as a natural language generation (NLG) task, such as machine translation \cite{10.1145/3406095}, in which models are trained on annotated datasets consisting of words and their corresponding explanations \cite{mickus-etal-2019-mark, gadetsky2018conditional}. Furthermore, as DM was born in the deep learning (DL) era, many DM approaches followed DL models, such as sequence-to-sequence architectures that require a myriad of annotated instances to train their weights properly. Hence, DM algorithms depend on the availability of large annotated datasets.  

Considering the importance of annotated data, numerous datasets have been established for the English \citelanguageresource{gadetsky2018conditional, mickus-etal-2019-mark, li-etal-2020-learning}. Recently, DM datasets have also been released for other languages, including Chinese \citelanguageresource{chang&chen2019}, French, German, Greek, Italian \citelanguageresource{kabiri2020evaluating} and Spanish \citelanguageresource{mickus2022codwoe}. The recent shared task, Semeval-2022 Task 1: CODWOE – Comparing Dictionaries and Word Embeddings \citelanguageresource{mickus2022codwoe}, has also contributed to the creation of other datasets. However, to the best of our knowledge, no DM dataset currently exists for Portuguese. In this research, we fill this gap by releasing \textsc{DORE}; the first dataset for \textbf{D}efinition M\textbf{O}delling in Po\textbf{R}tugu\textbf{E}se.

Portuguese is largely spoken officially on five continents, in seven countries, including Brazil and Portugal, and as a second language by more than 25 million people worldwide. Therefore, research in DM for Portuguese will be highly beneficial for millions of people, for which we lay the foundation through this paper by creating the first-ever Portuguese DM dataset, \textsc{DORE}. We also experiment with several DM methods on \textsc{DORE}. First, DM is performed as a sequence-to-sequence task using recent neural architectures. Then, we evaluate several popular large language models (LLMs), such as LLAMA2 and Falcon on Portuguese DM, using prompting. As they follow a zero-shot approach and do not need a training set, our findings can benefit a multitude of low-resource languages in definition modelling. As far as we know, this is the first time that LLMs are evaluated on low-resource DM.

Our \textbf{main contributions} can be summarised as follows:
\begin{enumerate}[label=\textcolor{blue}{(\arabic*)}] 
    \item We introduce \textsc{DORE}, the first dataset for Portuguese definition modelling, which comprises 103,019 definitions, and we describe the steps taken to compile it. 

    \item We evaluate several neural DM methods on \textsc{DORE} and report the results. 

    \item For the first time, we evaluate several popular LLMs on DM. We use prompting to generate definitions and compare the results.  
    

    \item We released \textsc{DORE}\footnote{\url{https://huggingface.co/datasets/multidefmod/dore}}, as an open-access dataset alongside the trained machine-learning models. 
\end{enumerate}

\section{Related Work}
\label{sec:related}
\paragraph{Datasets}
Definition Modelling (DM) has gained prominence as a deep learning problem, primarily due to its challenging nature. As mentioned before, most DM approaches have relied on supervised ML algorithms in which models are trained on annotated datasets. As a result, the NLP community has a growing interest in creating and collating datasets for DM. \citetlanguageresource{noraset2017} made available the first English dataset for the DM task, composed of definitions extracted from the Oxford Dictionary. Several English datasets were released in the following years. \citetlanguageresource{gadetsky2018conditional} and \citetlanguageresource{ZHANG2020113633} improved the dataset by \citetlanguageresource{noraset2017} by adding more instances. \citetlanguageresource{ishiwatari2019learning} released a DM dataset based on Wikipedia and Wikidata, 

\noindent DM datasets have been proposed to other languages as well. \citetlanguageresource{kabiri2020evaluating} released the first multilingual DM dataset, including Dutch, English, French, German, Greek, Italian, Japanese, Russian and Spanish. They utilised Wiktionary, OmegaWiki, and WordNet to extract the definitions. Furthermore, \citetlanguageresource{yang2020chineseDM} created the CDM dataset for the Chinese definition modelling task, where the definitions were extracted from the Chinese Concept Dictionary. As mentioned before,  Semeval-2022 Task 1 \citelanguageresource{mickus2022codwoe} also contributed to creating several DM datasets in several languages, including English, Spanish, French, Italian and Russian. \citetlanguageresource{huang-etal-2022-jade} further advances DM with a dataset for Japanese. However, as far as we know, there is no DM dataset available for Portuguese. 

\paragraph{Methods} In DM's introductory paper, \citet{noraset2017} presented an RNN-based model with an update function inspired by GRU gates to tackle word-to-sequence DM. The absence of relevant local contexts, however, hindered the production of definitions for polysemous words. To tackle this problem, \citet{gadetsky2018conditional} put forth two models that include contextual information for the first time. Later, \citet{ishiwatari2019learning} use local context (co-text) and global context (external information) to generate unknown definitions. They employ an LSTM-based encoder-decoder model and confirm that the generation task becomes harder when the words become more ambiguous and polysemous. Contrastively,  \citet{mickus-etal-2019-mark} recast DM as a sequence-to-sequence task rather than a word-to-sequence task; that is, context should be given as an input instead of the lemma.

Lately, transformer models revolutionised the NLP tasks \cite{devlin-etal-2019-bert}, and also DM. \citet{bevilacqua2020generationary} leverage BART \cite{lewis-etal-2020-bart} for tackling DM with Word-Sense Disambiguation and Word-in-Context tasks. Similarly, \citet{huang-etal-2021-definition} use a T5 \cite{raffel2020-t5-exploring} model to improve DM results significantly in several benchmarks. Finally, \citet{zhang-etal-2023-transling} explore generating bilingual definitions in English-Chinese by fine-tuning a pretrained multilingual machine translation model coupled with the exploitation of prompt combination and contrastive prompt learning. The model generates readable definitions but still produces hallucinations.

\vspace{-4mm}

\paragraph{CODWOE Shared Task \cite{mickus2022codwoe}} 
CODWOE focuses on generating glosses from vectors (DM track) and reconstructing embeddings from glosses (Reverse Dictionary track). They provided a second multilingual DM dataset, including English, French, Spanish, Italian, German, and Russian. Participants were encouraged to explore the potential benefits of multilingual and cross-lingual learning.

\begin{table*}[!ht]

\centering
\begin{tabular}{|l|c|c|c|c|}
    \hline
    \textbf{Dictionary} & \textbf{N. of Senses} & \textbf{Scraping} & \textbf{Context} & \textbf{Research use} \\
    \hline
    \href{https://michaelis.uol.com.br/moderno-portugues/busca/portugues-brasileiro/vari\%C3\%A2ncia/}{Dicionário Michaelis} & 350,000 & No & Partially & No \\
    \hline
    \href{https://houaiss.uol.com.br/corporativo/apps/uol_www/v6-0/html/index.php#0}{Dicionário Houaiss} & 376,500 & No & Partially & No \\
    \hline
    \href{https://aulete.com.br/}{Dicionário Aulete} & 818,000 & No & Partially & No \\
    \hline
    \href{https://dicionario.priberam.org/}{Dicionário Priberam} & 100,000 & No & Unordered examples & Yes \\
    \hline
    \href{https://support.google.com/websearch/answer/10106608?hl=pt-BR#zippy=}{Oxford Português} & 146,000 & Unknown & Partially & Yes \\
    \hline
    \href{https://www.dicio.com.br/}{Dicio} & 400,000 & Yes (Request) & Unordered examples & Yes \\
    \hline
    \href{https://pt.wiktionary.org/wiki/Wikcion\%C3\%A1rio:P\%C3\%A1gina_principal}{Portuguese Wiktionary} & > 270,501 & Yes & For some entries & Yes \\
    \hline    
\end{tabular}
\caption{Summary of potential data sources and their features}
\label{tab:datasources}
\end{table*}

\section{Dataset Construction}
\label{sec:data}
\subsection{Data Collection}
To collect data, we began by conducting extensive research into potential data sources, carefully evaluating their copyright status and quality. While DM typically relies on dictionary data, practical challenges in accessing these resources, as detailed in the following subsection, often make it necessary to leverage existing other resources. Subsequently, we extracted data from sources that aligned with our criteria.

\subsection{DORE dataset}
Definition Modelling relies on two primary resources: definitions and contexts, both typically found in dictionaries. Fortunately, recent technological advancements have made electronic dictionaries readily available, obviating the necessity for digitising printed materials. For the Portuguese language, surprisingly, e-dictionaries present unordered examples, which makes it challenging for readers (and machines) to connect them to corresponding senses.

Concerning the Portuguese, there are at least seven free monolingual e-dictionaries available for online consultation. They are: Michaelis, Houaiss, Aulete, Priberam, Portuguese Oxford (entries embedded into the Google search engine), Dicio and Portuguese Wiktionary.  
We survey these resources primarily because they are freely accessible and open to the public.

Table \ref{tab:datasources} summarises potential data sources and key features for this research, such as the number of senses, permission to scrape, research use permission, and the availability of the contexts. However, due to permission restrictions, we were only able to retrieve data from \textit{Dicio} and \textit{Portuguese Wiktionary}.

\vspace{-4mm}
\paragraph{Dicio \protect}
is a free e-dictionary that contains more than 400,000 senses. Entries include grammatical information (part of speech, plural form, etc.), definitions, and examples (occasionally). Dicio attempts to represent the contemporary Portuguese language and is conducive for research purposes.
\vspace{-4mm}
\paragraph{Wiktionary \protect}
is an online, crowdsourced dictionary aiming at becoming the universal polyglot dictionary. It covers more than 900 languages and features definitions, examples of use (occasionally), grammatical information (i.e., gender), and domain of use. For Portuguese, it contains more than 100,000 entries covering multiple varieties of the language.

To obtain data from the dictionaries, we employed a Python script to perform web scraping on each website. One notable challenge we encountered was the absence of comprehensive entry lists on these dictionary websites. Consequently, we resorted to employing word lists to generate the necessary URLs for data retrieval. The word lists used for creating the URLs in this dataset were sourced from \textit{Wiktionary} dumps provided by Kaikki's Project\footnote{\url{https://kaikki.org}} and word lists made available by Dicio. 

In Table \ref{tab:multidata}, we compare DORE with the other language resources available for definition modelling task in other languages. For the English dataset, we combined the data from the Oxford Dictionary \cite{gadetsky2018conditional}, the GCIDE and Wordnet dataset \cite{noraset2017}, the Wiktionary, Omega and Wordnet collected by \citet{kabiri2020evaluating}, and the CODWOE shared task \cite{mickus2022codwoe}. For the other languages, we combined data proposed by \citet{kabiri2020evaluating} and the CODWOE shared task \cite{mickus2022codwoe}. Although the results show that English boasts abundant resources for the DM task, featuring a vast number of instances, it is important to note that \textsc{DORE} has a compatible number of instances with other languages. Finallt, Table \ref{tab:dore_ex} shows examples of definitions from the \textsc{DORE} dataset with our respective translations.

\begin{table*}[!ht]
\centering
\begin{tabular}{|p{2cm}|p{13cm}|}
    \hline
    \textbf{Lemma} & \textbf{Definition}  \\
    \hline
   Abacaxi \newline \textcolor{blue}{(Pineapple)} & Planta originária do Brasil cultivada em muitas regiões quentes por causa de suas frutas de polpa açucarada e saborosa.  \newline \textcolor{blue}{(A plant native to Brazil, cultivated in many warm regions due to its sweet and tasty pulp.)} \\
   \hline
      Abacaxi \newline \textcolor{blue}{(Pineapple)} & [gíria] pessoa ou coisa maçante, complicada ou desagradável.  \newline \textcolor{blue}{([slang] a boring, unpleasant person or situation.)} \\
   \hline
      Florescência \newline \textcolor{blue}{(Flowering)} & [Botânica] Situação em que uma flor está no processo de maturação; antese.  \newline \textcolor{blue}{([Botany] Situation in which a flower is in the process of ripening; anthesis.)} \\
   \hline
    Florescência \newline \textcolor{blue}{(Taurean)} & [Figurado] Forte, como um touro.  \newline \textcolor{blue}{([Figurative] Strong, similar to a bull.)} \\
   \hline
   Taurino \newline \textcolor{blue}{(Flowering)} & Ação ou efeito de florescer; florescimento.  \newline \textcolor{blue}{(The act or effect of blooming; blossoming.)} \\
   \hline
   Desopilar \newline \textcolor{blue}{(Distract)} & [Figurado] Afastar da mente as preocupações ou os problemas; alegrar-se, divertir-se.  \newline \textcolor{blue}{([Figurative] Keep worries or problems out of the mind; rejoice, have fun.)} \\
   \hline
\end{tabular}
\caption{Instances of DORE dataset. English translations are in blue.}
\label{tab:dore_ex}
\end{table*}

\begin{table}[!ht]
\centering
\resizebox{\columnwidth}{!}{%
\begin{tabular}{|c|r|r|r|c|r|}
    \hline
    \textbf{Lag.} & \textbf{Lemmas} & \textbf{Unique} & \textbf{Senses} & \multicolumn{1}{|p{.2cm}|}{\textbf{Avg char.}} & \multicolumn{1}{|p{.2cm}|}{\textbf{Avg words}} \\
    \hline
    EN & 877,001 & 509,994 & 1.72 & 56.04 & 9.46 \\
    \hline
    FR & 200,880 & 55,068 & 6.2 & 75.63 & 14.30 \\
    \hline
    ES & 75,057 & 33,860 & 2.12 & 80.84 & 14.75 \\
    \hline
    IT & 62,465 & 35,987 & 1.73 & 78.97 & 13.61 \\
    \hline
    PT & 103,019& 27,978& 5.43& 72.38 & 11.38\\
    \hline
\end{tabular}
 }
 
\caption{Dataset statistics featuring language, number of instances, number of unique instances, number of senses per lemma, average number of characters per definition, and average number of words per definition, respectively.}
\label{tab:multidata}
\end{table}

\section{Methods}
\label{sec:methods}
In order to test the suitability of \textsc{DORE} for definition modelling, we exploit several deep learning models which are state-of-the-art in DM (Section \ref{sec:related}).

We first divided \textsc{DORE} into a training and test set following a 0.8 split of the complete dataset. Following machine learning models were used. We group models according to their architectures:

\vspace{-4mm}
\paragraph{General Transformers} -  We created a Seq2Seq model from general transformers by adding a transformer decoder, which takes the encoder's output and generates the target sequences. We only used the same transformer as the encoder and decoder. We experimented with several general-purpose transformer models that support Portuguese, including mBERT \cite{devlin-etal-2019-bert}, XLM-Roberta \cite{conneau-etal-2020-unsupervised}, and BERTimbau-large \cite{souza2020bertimbau}.

\vspace{-4mm}
\paragraph{Text Generation Transformers} - We also experimented with several text generation transformers as they have provided excellent results in English DM tasks. Specifically, we explored mBART \cite{lewis-etal-2020-bart} and several mT5 \cite{xue-etal-2021-mt5} variants. 

For both types of transformer models, we employed a batch size of 16, Adam optimiser with learning rate $1\mathrm{e}{-4}$, and a linear learning rate warm-up over 10\% of the training data. During the training process, the parameters of the transformer model were updated. The models were trained only using the training data and evaluated while training using an evaluation set that had one-fifth of the rows in training data. We performed early stopping if the evaluation loss did not improve over three evaluation steps. All the models were trained for three epochs. 

\vspace{-4mm}
\paragraph{LLMs} Finally, we evaluate how LLMs perform in \textsc{DORE}, a recent trend as we discussed before. We used two prompts to get a response from LLMs. For the instances where the context was available, we used the following prompt: "Provide the definition of \textit{\{WORD\}} appearing in this context  \textit{\{CONTEXT\}} in Portuguese".

For the instances where the context was not available, we used the following prompt: "Provide the definition of \textit{\{WORD\}}  in Portuguese."

We used several LLMs for prompting. We first use Davinci-003 through OpenAI API \cite{brown2020language}. Additionally, we used Falcon-7B-Instruct \cite{almazrouei2023falcon} and Llama-2-7B-32K-Instruct \cite{touvron2023llama}. All of these models are available in HuggingFace \cite{wolf-etal-2020-transformers}, and we use the LangChain implementation. As we followed a zero-shot prompting approach, we did not use any instances for the training set for LLMs. 

\section{Results}
\label{sec:results}

\begin{table}
\centering
\resizebox{\columnwidth}{!}{%
\begin{tabular}{|c|c|c|c|c|}
    \hline
    \textbf{Model} & \textbf{BLEU} & \textbf{TER} & \textbf{BLEURT} & \textbf{BERTScore} \\
    \hline
    \small mBERT & 0.18 & 0.78 & 0.52 &  0.61 \\
    XLM-R Large & 0.22 & 0.75 & 0.54 &  0.62 \\
    BERTimbau Large & 0.16 & 0.81 & 0.51 & 0.60 \\
    \hline
    mBART & 0.25 & 0.73 & 0.61 & 0.69  \\
    mT5 Base & 0.24 & 0.75 & 0.60 & 0.68  \\
    mT5 Large & 0.27 & 0.73 & 0.63 &  0.70 \\
    \hline
    GPT & 0.37 & 0.68 & \textbf{0.68} & \textbf{0.76}  \\
   Falcon 7B & 0.31 & 0.71 & 0.62 & 0.74  \\
    Llama 2 7B & 0.32 & 0.70 & 0.64  & 0.72 \\
    \hline
\end{tabular}
}
\caption{The result of different ML models in DORE test set by the different ML architectures}
\label{tab:results}
\end{table}

The results of the aforementioned models are shown in Table \ref{tab:results}. All the models were evaluated using the test set. We used several evaluation metrics to compare the models, BLEU \cite{papineniBLEU2002}, and TER \cite{snover-etal-2006-study}. However, both of these metrics lack semantic understanding. Therefore, we also used two recent NLG evaluation metrics, BLEURT \cite{sellam-etal-2020-bleurt} and BERTScore \cite{zhang2019bertscore}. 

BLEU and TER, commonly known as NLG metrics, report low results for all groups, which resonates with previous NLG and DM investigations in that current NLG metrics are not satisfactory \cite{mickus2022codwoe}.

Unsurprisingly, the best performing group is the LLMs, probably due to their incomparable parameter size and pre-embedded encyclopedic knowledge. GPT outperforms other LLMs slightly. It is worth noting that text generation transformers closely trail the LLM results, with the mT5 Large model surpassing its base variant and mBART. These numbers are compatible with those obtained in experiments with smaller datasets in other romance languages, such as French, Spanish and Italian \cite{mickus2022codwoe}. However, BERTimbau Large model, which is explicitly trained on Portuguese text, provides the worst results from the experimented models. Overall, the results demonstrate that language models designed explicitly for text generation excel in the DM task in Portuguese, even when they are multilingual.

\section{Conclusion}
\label{sec:conclusion}
We introduced DORE, the first dataset for automatic generation of definitions in Portuguese. We demonstrate DORE's usefulness by performing Definition Modelling for Portuguese for the first time with several pretrained models together with popular LLMs. The results show that LLMs perform better in Portuguese DM. We released DORE and our code publicly with a view to fostering more research on various tasks in Portuguese. 

As future work, we intend to expand DORE with more instances of definitions. We also plan to include in-context examples of lemmas, which can be useful for future experiments and other NLP tasks, such as word sense disambiguation and word in context. Besides that, we also plan to harness other datasets to perform cross-lingual learning.

\section*{Acknowledgments} 

The computational experiments in this paper were conducted on the Aston EPS Machine Learning Server, funded by the EPSRC Core Equipment Fund, Grant EP/V036106/1.

\section*{Ethics Statement}
As mentioned in \ref{sec:data}, the Data section, DORE was collected from publicly available resources, and none of the definitions were edited. We sought permission from Dicio to use definitions for this research. Similar to previous research, we shared the definitions and their lemmas. Also, we released DORE and corresponding models under the Creative Commons Attribution-Non Commercial-ShareAlike (CC-BY-NC-SA) 4.0 International Public License, which prevents users from editing any instances of the dataset. While DORE and related models are publicly available, we released it as a gated dataset so that users need to comply with the license to request access. We reinforce that models and dataset should be used for research only.

\nocite{*}
\section{Bibliographical References}\label{sec:reference}

\bibliographystyle{lrec-coling2024-natbib}
\bibliography{lrec-coling2024-example}

\begin{thebibliography}{11}
\expandafter\ifx\csname natexlab\endcsname\relax\def\natexlab#1{#1}\fi

\bibitem[{Chang and Chen(2019)}]{chang&chen2019}
Ting-Yun Chang and Yun-Nung Chen. 2019.
\newblock What does this word mean? explaining contextualized embeddings with
  natural language definition.
\newblock In \emph{Proceedings of the 2019 Conference on Empirical Methods in
  Natural Language Processing and the 9th International Joint Conference on
  Natural Language Processing (EMNLP-IJCNLP)}, pages 6064--6070.

\bibitem[{Gadetsky et~al.(2018)Gadetsky, Yakubovskiy, and
  Vetrov}]{gadetsky2018conditional}
Artyom Gadetsky, Ilya Yakubovskiy, and Dmitry Vetrov. 2018.
\newblock \href {https://doi.org/10.18653/v1/P18-2043} {Conditional generators
  of words definitions}.
\newblock In \emph{Proceedings of the 56th Annual Meeting of the Association
  for Computational Linguistics (Volume 2: Short Papers)}, pages 266--271,
  Melbourne, Australia. Association for Computational Linguistics.

\bibitem[{Huang et~al.(2022)Huang, Kajiwara, and Arase}]{huang-etal-2022-jade}
Han Huang, Tomoyuki Kajiwara, and Yuki Arase. 2022.
\newblock \href {https://aclanthology.org/2022.lrec-1.743} {{JADE}: Corpus for
  {J}apanese definition modelling}.
\newblock In \emph{Proceedings of the Thirteenth Language Resources and
  Evaluation Conference}, pages 6884--6888, Marseille, France. European
  Language Resources Association.

\bibitem[{Ishiwatari et~al.(2019)Ishiwatari, Hayashi, Yoshinaga, Neubig, Sato,
  Toyoda, and Kitsuregawa}]{ishiwatari2019learning}
Shonosuke Ishiwatari, Hiroaki Hayashi, Naoki Yoshinaga, Graham Neubig, Shoetsu
  Sato, Masashi Toyoda, and Masaru Kitsuregawa. 2019.
\newblock Learning to describe unknown phrases with local and global contexts.
\newblock In \emph{Proceedings of the 2019 Conference of the North American
  Chapter of the Association for Computational Linguistics: Human Language
  Technologies, Volume 1 (Long and Short Papers)}, pages 3467--3476.

\bibitem[{Kabiri and Cook(2020)}]{kabiri2020evaluating}
Arman Kabiri and Paul Cook. 2020.
\newblock Evaluating a multi-sense definition generation model for multiple
  languages.
\newblock In \emph{International Conference on Text, Speech, and Dialogue},
  pages 153--161. Springer.

\bibitem[{Li et~al.(2020)Li, Hu, Zhang, Xu, Jiang, Xiao, Zhu, Liu, and
  Li}]{li-etal-2020-learning}
Yinqiao Li, Chi Hu, Yuhao Zhang, Nuo Xu, Yufan Jiang, Tong Xiao, Jingbo Zhu,
  Tongran Liu, and Changliang Li. 2020.
\newblock \href {https://doi.org/10.18653/v1/2020.acl-main.592} {Learning
  architectures from an extended search space for language modeling}.
\newblock In \emph{Proceedings of the 58th Annual Meeting of the Association
  for Computational Linguistics}, pages 6629--6639, Online. Association for
  Computational Linguistics.

\bibitem[{Mickus et~al.(2019)Mickus, Paperno, and
  Constant}]{mickus-etal-2019-mark}
Timothee Mickus, Denis Paperno, and Matthieu Constant. 2019.
\newblock \href {https://aclanthology.org/W19-6201} {Mark my word: A
  sequence-to-sequence approach to definition modeling}.
\newblock In \emph{Proceedings of the First NLPL Workshop on Deep Learning for
  Natural Language Processing}, pages 1--11, Turku, Finland. Link{\"o}ping
  University Electronic Press.

\bibitem[{Mickus et~al.(2022)Mickus, Van~Deemter, Constant, and
  Paperno}]{mickus2022codwoe}
Timothee Mickus, Kees Van~Deemter, Mathieu Constant, and Denis Paperno. 2022.
\newblock \href {https://doi.org/10.18653/v1/2022.semeval-1.1} {{S}emeval-2022
  task 1: {CODWOE} {--} comparing dictionaries and word embeddings}.
\newblock In \emph{Proceedings of the 16th International Workshop on Semantic
  Evaluation (SemEval-2022)}, pages 1--14, Seattle, United States. Association
  for Computational Linguistics.

\bibitem[{Noraset et~al.(2017)Noraset, Liang, Birnbaum, and
  Downey}]{noraset2017}
Thanapon Noraset, Chen Liang, Larry Birnbaum, and Doug Downey. 2017.
\newblock Definition modeling: Learning to define word embeddings in natural
  language.
\newblock In \emph{Proceedings of the Thirty-First AAAI Conference on
  Artificial Intelligence}, AAAI'17, page 3259–3266. AAAI Press.

\bibitem[{Yang et~al.(2020)Yang, Kong, Chen, Liu, Fan, and
  Yang}]{yang2020chineseDM}
Liner Yang, Cunliang Kong, Yun Chen, Yang Liu, Qinan Fan, and Erhong Yang.
  2020.
\newblock Incorporating sememes into chinese definition modeling.
\newblock \emph{IEEE/ACM Transactions on Audio, Speech, and Language
  Processing}, 28:1669--1677.

\bibitem[{Zhang et~al.(2020)Zhang, Du, Sun, and Li}]{ZHANG2020113633}
Haitong Zhang, Yongping Du, Jiaxin Sun, and Qingxiao Li. 2020.
\newblock \href {https://doi.org/https://doi.org/10.1016/j.eswa.2020.113633}
  {Improving interpretability of word embeddings by generating definition and
  usage}.
\newblock \emph{Expert Systems with Applications}, 160:113633.

\end{thebibliography}


\begin{thebibliography}{32}
\expandafter\ifx\csname natexlab\endcsname\relax\def\natexlab#1{#1}\fi

\bibitem[{Almazrouei et~al.(2023)Almazrouei, Alobeidli, Alshamsi, Cappelli,
  Cojocaru, Debbah, Goffinet, Hesslow, Launay, Malartic
  et~al.}]{almazrouei2023falcon}
Ebtesam Almazrouei, Hamza Alobeidli, Abdulaziz Alshamsi, Alessandro Cappelli,
  Ruxandra Cojocaru, M{\'e}rouane Debbah, {\'E}tienne Goffinet, Daniel Hesslow,
  Julien Launay, Quentin Malartic, et~al. 2023.
\newblock The falcon series of open language models.
\newblock \emph{arXiv preprint arXiv:2311.16867}.

\bibitem[{Bevilacqua et~al.(2020)Bevilacqua, Maru, and
  Navigli}]{bevilacqua2020generationary}
Michele Bevilacqua, Marco Maru, and Roberto Navigli. 2020.
\newblock Generationary or “how we went beyond word sense inventories and
  learned to gloss”.
\newblock In \emph{Proceedings of the 2020 Conference on Empirical Methods in
  Natural Language Processing (EMNLP)}, pages 7207--7221.

\bibitem[{Brown et~al.(2020)Brown, Mann, Ryder, Subbiah, Kaplan, Dhariwal,
  Neelakantan, Shyam, Sastry, Askell et~al.}]{brown2020language}
Tom Brown, Benjamin Mann, Nick Ryder, Melanie Subbiah, Jared~D Kaplan, Prafulla
  Dhariwal, Arvind Neelakantan, Pranav Shyam, Girish Sastry, Amanda Askell,
  et~al. 2020.
\newblock Language models are few-shot learners.
\newblock \emph{Advances in neural information processing systems},
  33:1877--1901.

\bibitem[{Chang and Chen(2019)}]{chang&chen2019}
Ting-Yun Chang and Yun-Nung Chen. 2019.
\newblock What does this word mean? explaining contextualized embeddings with
  natural language definition.
\newblock In \emph{Proceedings of the 2019 Conference on Empirical Methods in
  Natural Language Processing and the 9th International Joint Conference on
  Natural Language Processing (EMNLP-IJCNLP)}, pages 6064--6070.

\bibitem[{Conneau et~al.(2020)Conneau, Khandelwal, Goyal, Chaudhary, Wenzek,
  Guzm{\'a}n, Grave, Ott, Zettlemoyer, and
  Stoyanov}]{conneau-etal-2020-unsupervised}
Alexis Conneau, Kartikay Khandelwal, Naman Goyal, Vishrav Chaudhary, Guillaume
  Wenzek, Francisco Guzm{\'a}n, Edouard Grave, Myle Ott, Luke Zettlemoyer, and
  Veselin Stoyanov. 2020.
\newblock \href {https://doi.org/10.18653/v1/2020.acl-main.747} {Unsupervised
  cross-lingual representation learning at scale}.
\newblock In \emph{Proceedings of the 58th Annual Meeting of the Association
  for Computational Linguistics}, pages 8440--8451, Online. Association for
  Computational Linguistics.

\bibitem[{Dabre et~al.(2020)Dabre, Chu, and Kunchukuttan}]{10.1145/3406095}
Raj Dabre, Chenhui Chu, and Anoop Kunchukuttan. 2020.
\newblock \href {https://doi.org/10.1145/3406095} {A survey of multilingual
  neural machine translation}.
\newblock \emph{ACM Comput. Surv.}, 53(5).

\bibitem[{Devlin et~al.(2019)Devlin, Chang, Lee, and
  Toutanova}]{devlin-etal-2019-bert}
Jacob Devlin, Ming-Wei Chang, Kenton Lee, and Kristina Toutanova. 2019.
\newblock \href {https://doi.org/10.18653/v1/N19-1423} {{BERT}: Pre-training of
  deep bidirectional transformers for language understanding}.
\newblock In \emph{Proceedings of the 2019 Conference of the North {A}merican
  Chapter of the Association for Computational Linguistics: Human Language
  Technologies, Volume 1 (Long and Short Papers)}, pages 4171--4186,
  Minneapolis, Minnesota. Association for Computational Linguistics.

\bibitem[{Domínguez~Vázquez and Gouws(2023)}]{vazquez-gouws-2023}
María~José Domínguez~Vázquez and Rufus~H Gouws. 2023.
\newblock \href {https://doi.org/10.1093/ijl/ecac020} {{The Definition,
  Presentation and Automatic Generation of Contextual Data in Lexicography}}.
\newblock \emph{International Journal of Lexicography}, page ecac020.

\bibitem[{Dziemianko(2020)}]{dziemianko-2020-dict-usefulness}
Anna Dziemianko. 2020.
\newblock \href {https://doi.org/10.1093/ijl/ecaa017} {{Smart advertising and
  online dictionary usefulness}}.
\newblock \emph{International Journal of Lexicography}, 33(4):377--403.

\bibitem[{Gadetsky et~al.(2018)Gadetsky, Yakubovskiy, and
  Vetrov}]{gadetsky2018conditional}
Artyom Gadetsky, Ilya Yakubovskiy, and Dmitry Vetrov. 2018.
\newblock \href {https://doi.org/10.18653/v1/P18-2043} {Conditional generators
  of words definitions}.
\newblock In \emph{Proceedings of the 56th Annual Meeting of the Association
  for Computational Linguistics (Volume 2: Short Papers)}, pages 266--271,
  Melbourne, Australia. Association for Computational Linguistics.

\bibitem[{Huang et~al.(2021)Huang, Kajiwara, and
  Arase}]{huang-etal-2021-definition}
Han Huang, Tomoyuki Kajiwara, and Yuki Arase. 2021.
\newblock \href {https://doi.org/10.18653/v1/2021.emnlp-main.194} {Definition
  modelling for appropriate specificity}.
\newblock In \emph{Proceedings of the 2021 Conference on Empirical Methods in
  Natural Language Processing}, pages 2499--2509, Online and Punta Cana,
  Dominican Republic. Association for Computational Linguistics.

\bibitem[{Ishiwatari et~al.(2019)Ishiwatari, Hayashi, Yoshinaga, Neubig, Sato,
  Toyoda, and Kitsuregawa}]{ishiwatari2019learning}
Shonosuke Ishiwatari, Hiroaki Hayashi, Naoki Yoshinaga, Graham Neubig, Shoetsu
  Sato, Masashi Toyoda, and Masaru Kitsuregawa. 2019.
\newblock Learning to describe unknown phrases with local and global contexts.
\newblock In \emph{Proceedings of the 2019 Conference of the North American
  Chapter of the Association for Computational Linguistics: Human Language
  Technologies, Volume 1 (Long and Short Papers)}, pages 3467--3476.

\bibitem[{Kabiri and Cook(2020)}]{kabiri2020evaluating}
Arman Kabiri and Paul Cook. 2020.
\newblock Evaluating a multi-sense definition generation model for multiple
  languages.
\newblock In \emph{International Conference on Text, Speech, and Dialogue},
  pages 153--161. Springer.

\bibitem[{Lewis et~al.(2020)Lewis, Liu, Goyal, Ghazvininejad, Mohamed, Levy,
  Stoyanov, and Zettlemoyer}]{lewis-etal-2020-bart}
Mike Lewis, Yinhan Liu, Naman Goyal, Marjan Ghazvininejad, Abdelrahman Mohamed,
  Omer Levy, Veselin Stoyanov, and Luke Zettlemoyer. 2020.
\newblock \href {https://doi.org/10.18653/v1/2020.acl-main.703} {{BART}:
  Denoising sequence-to-sequence pre-training for natural language generation,
  translation, and comprehension}.
\newblock In \emph{Proceedings of the 58th Annual Meeting of the Association
  for Computational Linguistics}, pages 7871--7880, Online. Association for
  Computational Linguistics.

\bibitem[{Li et~al.(2020)Li, Hu, Zhang, Xu, Jiang, Xiao, Zhu, Liu, and
  Li}]{li-etal-2020-learning}
Yinqiao Li, Chi Hu, Yuhao Zhang, Nuo Xu, Yufan Jiang, Tong Xiao, Jingbo Zhu,
  Tongran Liu, and Changliang Li. 2020.
\newblock \href {https://doi.org/10.18653/v1/2020.acl-main.592} {Learning
  architectures from an extended search space for language modeling}.
\newblock In \emph{Proceedings of the 58th Annual Meeting of the Association
  for Computational Linguistics}, pages 6629--6639, Online. Association for
  Computational Linguistics.

\bibitem[{Mickus et~al.(2019)Mickus, Paperno, and
  Constant}]{mickus-etal-2019-mark}
Timothee Mickus, Denis Paperno, and Matthieu Constant. 2019.
\newblock \href {https://aclanthology.org/W19-6201} {Mark my word: A
  sequence-to-sequence approach to definition modeling}.
\newblock In \emph{Proceedings of the First NLPL Workshop on Deep Learning for
  Natural Language Processing}, pages 1--11, Turku, Finland. Link{\"o}ping
  University Electronic Press.

\bibitem[{Mickus et~al.(2022)Mickus, Van~Deemter, Constant, and
  Paperno}]{mickus2022codwoe}
Timothee Mickus, Kees Van~Deemter, Mathieu Constant, and Denis Paperno. 2022.
\newblock \href {https://doi.org/10.18653/v1/2022.semeval-1.1} {{S}emeval-2022
  task 1: {CODWOE} {--} comparing dictionaries and word embeddings}.
\newblock In \emph{Proceedings of the 16th International Workshop on Semantic
  Evaluation (SemEval-2022)}, pages 1--14, Seattle, United States. Association
  for Computational Linguistics.

\bibitem[{Ni and Wang(2017)}]{ni-wang-2017-learning}
Ke~Ni and William~Yang Wang. 2017.
\newblock \href {https://aclanthology.org/I17-2070} {Learning to explain
  non-standard {E}nglish words and phrases}.
\newblock In \emph{Proceedings of the Eighth International Joint Conference on
  Natural Language Processing (Volume 2: Short Papers)}, pages 413--417,
  Taipei, Taiwan. Asian Federation of Natural Language Processing.

\bibitem[{Noraset et~al.(2017)Noraset, Liang, Birnbaum, and
  Downey}]{noraset2017}
Thanapon Noraset, Chen Liang, Larry Birnbaum, and Doug Downey. 2017.
\newblock Definition modeling: Learning to define word embeddings in natural
  language.
\newblock In \emph{Proceedings of the Thirty-First AAAI Conference on
  Artificial Intelligence}, AAAI'17, page 3259–3266. AAAI Press.

\bibitem[{Papineni et~al.(2002)Papineni, Roukos, Ward, and
  Zhu}]{papineniBLEU2002}
Kishore Papineni, Salim Roukos, Todd Ward, and Wei-Jing Zhu. 2002.
\newblock \href {https://doi.org/10.3115/1073083.1073135} {Bleu: A method for
  automatic evaluation of machine translation}.
\newblock In \emph{Proceedings of the 40th Annual Meeting on Association for
  Computational Linguistics}, ACL '02, page 311–318, USA. Association for
  Computational Linguistics.

\bibitem[{Raffel et~al.(2020)Raffel, Shazeer, Roberts, Lee, Narang, Matena,
  Zhou, Li, and Liu}]{raffel2020-t5-exploring}
Colin Raffel, Noam Shazeer, Adam Roberts, Katherine Lee, Sharan Narang, Michael
  Matena, Yanqi Zhou, Wei Li, and Peter~J Liu. 2020.
\newblock Exploring the limits of transfer learning with a unified text-to-text
  transformer.
\newblock \emph{Journal of Machine Learning Research}, 21:1--67.

\bibitem[{San~Martín(2021)}]{sanmartin-2023-termdefs}
Antonio San~Martín. 2021.
\newblock \href {https://doi.org/10.1093/ijl/ecab013} {{A Flexible Approach to
  Terminological Definitions: Representing Thematic Variation}}.
\newblock \emph{International Journal of Lexicography}, 35(1):53--74.

\bibitem[{Sellam et~al.(2020)Sellam, Das, and Parikh}]{sellam-etal-2020-bleurt}
Thibault Sellam, Dipanjan Das, and Ankur Parikh. 2020.
\newblock \href {https://doi.org/10.18653/v1/2020.acl-main.704} {{BLEURT}:
  Learning robust metrics for text generation}.
\newblock In \emph{Proceedings of the 58th Annual Meeting of the Association
  for Computational Linguistics}, pages 7881--7892, Online. Association for
  Computational Linguistics.

\bibitem[{Snover et~al.(2006)Snover, Dorr, Schwartz, Micciulla, and
  Makhoul}]{snover-etal-2006-study}
Matthew Snover, Bonnie Dorr, Rich Schwartz, Linnea Micciulla, and John Makhoul.
  2006.
\newblock \href {https://aclanthology.org/2006.amta-papers.25} {A study of
  translation edit rate with targeted human annotation}.
\newblock In \emph{Proceedings of the 7th Conference of the Association for
  Machine Translation in the Americas: Technical Papers}, pages 223--231,
  Cambridge, Massachusetts, USA. Association for Machine Translation in the
  Americas.

\bibitem[{Souza et~al.(2020)Souza, Nogueira, and Lotufo}]{souza2020bertimbau}
F{\'a}bio Souza, Rodrigo Nogueira, and Roberto Lotufo. 2020.
\newblock Bertimbau: pretrained bert models for brazilian portuguese.
\newblock In \emph{Intelligent Systems: 9th Brazilian Conference, BRACIS 2020,
  Rio Grande, Brazil, October 20--23, 2020, Proceedings, Part I 9}, pages
  403--417. Springer.

\bibitem[{Touvron et~al.(2023)Touvron, Martin, Stone, Albert, Almahairi,
  Babaei, Bashlykov, Batra, Bhargava, Bhosale et~al.}]{touvron2023llama}
Hugo Touvron, Louis Martin, Kevin Stone, Peter Albert, Amjad Almahairi, Yasmine
  Babaei, Nikolay Bashlykov, Soumya Batra, Prajjwal Bhargava, Shruti Bhosale,
  et~al. 2023.
\newblock Llama 2: Open foundation and fine-tuned chat models.
\newblock \emph{arXiv preprint arXiv:2307.09288}.

\bibitem[{Wolf et~al.(2020)Wolf, Debut, Sanh, Chaumond, Delangue, Moi, Cistac,
  Rault, Louf, Funtowicz, Davison, Shleifer, von Platen, Ma, Jernite, Plu, Xu,
  Le~Scao, Gugger, Drame, Lhoest, and Rush}]{wolf-etal-2020-transformers}
Thomas Wolf, Lysandre Debut, Victor Sanh, Julien Chaumond, Clement Delangue,
  Anthony Moi, Pierric Cistac, Tim Rault, Remi Louf, Morgan Funtowicz, Joe
  Davison, Sam Shleifer, Patrick von Platen, Clara Ma, Yacine Jernite, Julien
  Plu, Canwen Xu, Teven Le~Scao, Sylvain Gugger, Mariama Drame, Quentin Lhoest,
  and Alexander Rush. 2020.
\newblock \href {https://doi.org/10.18653/v1/2020.emnlp-demos.6} {Transformers:
  State-of-the-art natural language processing}.
\newblock In \emph{Proceedings of the 2020 Conference on Empirical Methods in
  Natural Language Processing: System Demonstrations}, pages 38--45, Online.
  Association for Computational Linguistics.

\bibitem[{Xue et~al.(2021)Xue, Constant, Roberts, Kale, Al-Rfou, Siddhant,
  Barua, and Raffel}]{xue-etal-2021-mt5}
Linting Xue, Noah Constant, Adam Roberts, Mihir Kale, Rami Al-Rfou, Aditya
  Siddhant, Aditya Barua, and Colin Raffel. 2021.
\newblock \href {https://doi.org/10.18653/v1/2021.naacl-main.41} {m{T}5: A
  massively multilingual pre-trained text-to-text transformer}.
\newblock In \emph{Proceedings of the 2021 Conference of the North American
  Chapter of the Association for Computational Linguistics: Human Language
  Technologies}, pages 483--498, Online. Association for Computational
  Linguistics.

\bibitem[{Yang et~al.(2020)Yang, Kong, Chen, Liu, Fan, and
  Yang}]{yang2020chineseDM}
Liner Yang, Cunliang Kong, Yun Chen, Yang Liu, Qinan Fan, and Erhong Yang.
  2020.
\newblock Incorporating sememes into chinese definition modeling.
\newblock \emph{IEEE/ACM Transactions on Audio, Speech, and Language
  Processing}, 28:1669--1677.

\bibitem[{Zhang et~al.(2020)Zhang, Du, Sun, and Li}]{ZHANG2020113633}
Haitong Zhang, Yongping Du, Jiaxin Sun, and Qingxiao Li. 2020.
\newblock \href {https://doi.org/https://doi.org/10.1016/j.eswa.2020.113633}
  {Improving interpretability of word embeddings by generating definition and
  usage}.
\newblock \emph{Expert Systems with Applications}, 160:113633.

\bibitem[{Zhang et~al.(2023)Zhang, Li, Li, Shang, Shi, and
  Jiang}]{zhang-etal-2023-transling}
Hengyuan Zhang, Dawei Li, Yanran Li, Chenming Shang, Chufan Shi, and Yong
  Jiang. 2023.
\newblock \href {https://doi.org/10.18653/v1/2023.bea-1.23} {Assisting language
  learners: Automated trans-lingual definition generation via contrastive
  prompt learning}.
\newblock In \emph{Proceedings of the 18th Workshop on Innovative Use of NLP
  for Building Educational Applications (BEA 2023)}, pages 260--274, Toronto,
  Canada. Association for Computational Linguistics.

\bibitem[{Zhang et~al.(2019)Zhang, Kishore, Wu, Weinberger, and
  Artzi}]{zhang2019bertscore}
Tianyi Zhang, Varsha Kishore, Felix Wu, Kilian~Q Weinberger, and Yoav Artzi.
  2019.
\newblock Bertscore: Evaluating text generation with bert.
\newblock In \emph{International Conference on Learning Representations}.

\end{thebibliography}

\section{Language Resource References}
\label{lr:ref}
\bibliographystylelanguageresource{lrec-coling2024-natbib}
\bibliographylanguageresource{languageresource}

\newpage
\end{document}